\pdfoutput=1
\documentclass[11pt]{article}
\usepackage{acl}
\usepackage{times}
\usepackage{latexsym}
\usepackage[T1]{fontenc}
\usepackage[utf8]{inputenc}
\usepackage{microtype}
\usepackage{inconsolata}
\usepackage{graphicx}
\usepackage{amsmath} 
\usepackage{amsfonts}
\usepackage{float}
\usepackage{booktabs}
\usepackage{xspace}
\usepackage{tabularx}  
\usepackage{dblfloatfix}
\usepackage{bm}

\title{MindFlow: Revolutionizing E-commerce Customer Support with Multimodal LLM Agents}
\author{
Ming Gong\textsuperscript{\rm1,\rm2}, 
Xucheng Huang\textsuperscript{\rm1}, 
Chenghan Yang\textsuperscript{\rm1}, 
Xianhan Peng\textsuperscript{\rm1}, 
Haoxin Wang\textsuperscript{\rm1}, 
Yang Liu\textsuperscript{\rm1}, 
Ling Jiang\textsuperscript{\rm1}\\
\textsuperscript{\rm1}Xiaoduo AI,
\textsuperscript{\rm2}University of Dayton \\
\texttt{gongm1@udayton.edu,
liuyangfoam@xiaoduotech.com}
}

\begin{document}
\maketitle
\begin{abstract}
Recent advances in large language models (LLMs) have enabled new applications in e-commerce customer service. However, their capabilities remain constrained in complex, multimodal scenarios. We present \textbf{MindFlow}, the first open-source multimodal LLM agent tailored for e-commerce. Built on the CoALA framework, it integrates memory, decision-making, and action modules, and adopts a modular “MLLM-as-Tool” strategy for efficient visual-textual reasoning. Evaluated via online A/B testing and simulation-based ablation, MindFlow demonstrates substantial gains in handling complex queries, improving user satisfaction, and reducing operational costs, with a \textbf{93.53\%} relative improvement observed in real-world deployments. The code will be released upon publication to support future research.
\end{abstract}
\section{Introduction}

The exponential growth of e-commerce has placed increasing demands on customer service systems , which are now expected to resolve complex multimodal inquiries in real time while maintaining high user satisfaction~\citep{gajewska2020impact}. Despite recent progress in LLMs, significant challenges remain in real-world deployment~\citep{chaturvedi2023opportunities,ren2024survey}. Current systems commonly struggle with three key challenges: integrating visual and textual inputs effectively, maintaining contextual coherence across multi-turn dialogues, and making adaptive decisions in dynamic, open-ended scenarios. These shortcomings collectively result in lower resolution rates, higher operational costs, and degraded user experience.

To address these challenges, we propose MindFlow, a novel multimodal LLM agent framework designed specifically for e-commerce customer service. Built on the CoALA architecture~\cite{sumers2023cognitive}, MindFlow tightly integrates a memory module, a decision-making module and an action module to generate precise, context-aware responses. It further adopts the ``MLLM-as-Tool'' paradigm, which decomposes visual-semantic tasks and leverages Multimodal Large Language Models (MLLMs) for more efficient visual-textual reasoning in complex multimodal interactions.


We conduct comprehensive evaluations through online A/B testing and simulation-based ablation studies. Real-world online A/B testing demonstrates significant gains over traditional rule-based systems, with reduced latency and operational costs. Together with extensive simulation-based ablation studies, these results underscore the effectiveness of our decision-making and action modules, and highlight the robustness of the ``MLLM-as-Tool'' paradigm for multimodal reasoning in practical deployments.

Our primary contributions are:

\begin{itemize}
    \item \textbf{First open-source multimodal LLM agent for industrial e-commerce}, providing a practical framework with tightly integrated memory, decision-making, and action modules for real-time, context-aware reasoning.
    \item \textbf{Modular “MLLM-as-Tool” paradigm}, introducing a novel strategy that treats MLLMs as specialized visual processing tools to efficiently decompose and solve complex visual-textual tasks.
    \item \textbf{Extensive experimental validation}, including online A/B testing and simulation-based ablation studies, confirming MindFlow’s effectiveness, robustness, and scalability in realistic e-commerce scenarios.
\end{itemize}

\section{Related Work}

Language agents are an emerging class of AI systems that integrate LLMs with external tools, memory modules, and environment interfaces to enable autonomous reasoning and decision-making~\citep{xi2023rise, weng2023agent, wang2024survey}. Unlike static prompting, these agents operate in a closed-loop fashion: they formulate actions, receive feedback, and iteratively update internal state.

Influential methods such as ReAct~\citep{yao2023react} combine step-by-step reasoning with tool use, allowing agents to interleave thoughts and actions for enhanced problem-solving. Reflexion~\citep{shinn2023reflexion} further introduces self-evaluation and retrospective learning, enabling iterative strategy refinement. Building upon these foundations, recent systems like CAMEL~\citep{li2023camel} support multi-agent collaboration, whereas SWE-agent~\citep{yang2024swe} specializes in domain-specific workflows. Meanwhile, autonomous agents such as Manus and BabyAGI~\citep{yoheinakajima2023babyagi} investigate open-ended task decomposition and planning in dynamic environments.

In the domain of e-commerce customer service, LLM agents are increasingly explored to blend automated assistance with human-in-the-loop workflows. While public documentation on commercial deployments remains limited, systems such as Crescendo, Sierra AI, and Alibaba’s 1688 platform have reportedly integrated AI agents for recommendation, dialogue, and post-sale support. Complementing these applications, recent research proposes instruction-tuned and domain-adapted models~\citep{herold2024lilium,peng2024ecellm,zhang2024llasa}, leveraging structured data and multitask learning to tackle challenges like product understanding and customer intent classification. These developments underscore the growing emergence of LLM agents in e-commerce applications.

\section{MindFlow}

\begin{figure*}[!htb]
    \centering
    \includegraphics[width=1.0\textwidth]{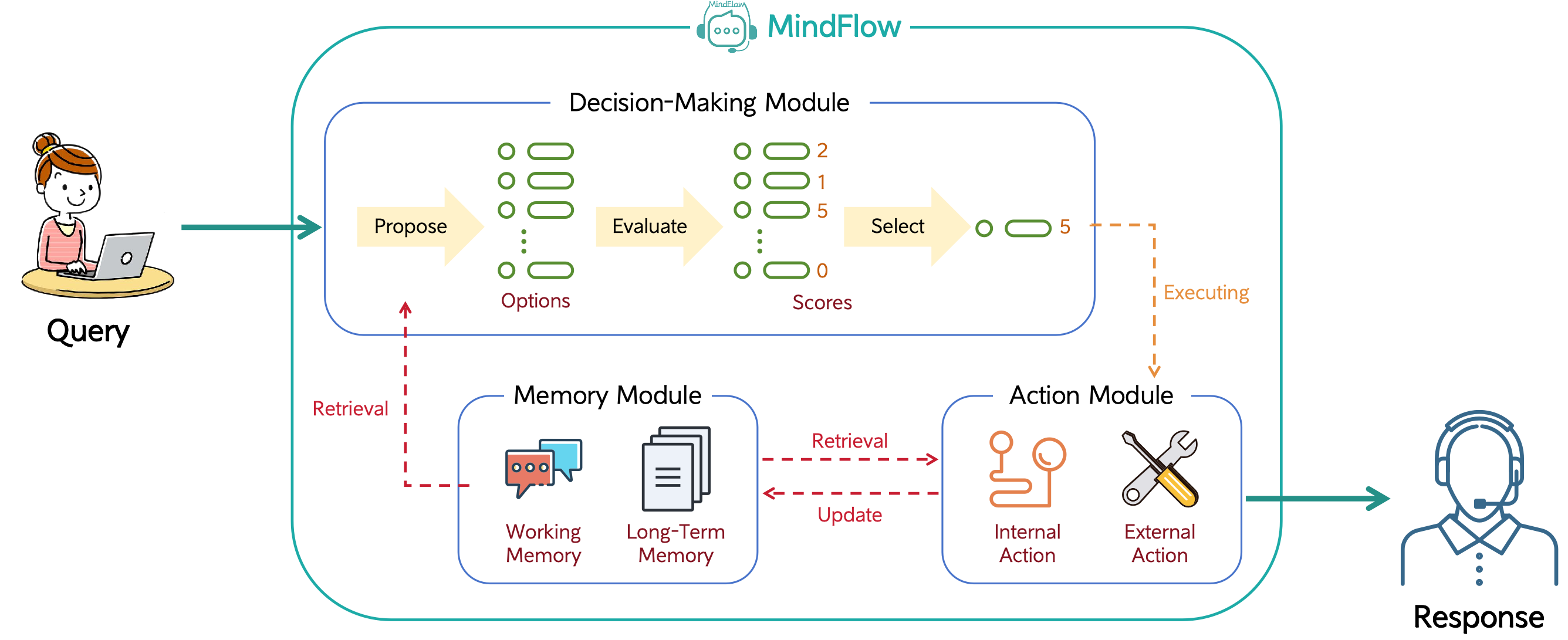}
    \caption{MindFlow Architecture}
    \label{fig:arch}
\end{figure*}

MindFlow, implemented using OpenAI Agent SDK, functions as an agentic agent capable of invoking external tools encapsulated in the Model Context Protocol (MCP) format. Figure~\ref{fig:arch} presents the overall architecture of MindFlow, which processes buyer queries through a tightly integrated loop of decision-making, memory, and action modules to generate reliable and context-aware responses. Its capability to handle multimodal scenarios is further enhanced by the integration of the ``MLLM-as-Tool'' strategy.

\subsection{Memory Module}
The memory module is introduced to enhance contextual understanding and knowledge retention in complex e-commerce tasks. It consists of two complementary components: working memory and long-term memory.

The working memory captures the historical interaction between the buyer and the customer representative. It is accessed by the decision-making module during the Propose stage to help infer buyer intent, particularly when queries are implicit or incomplete. In contrast, the long-term memory stores domain-specific knowledge, including platform policies, store-level promotions, product details, and buyer-specific data such as order and logistics information. This component is retrieved through internal actions in the action module and is continuously updated as new information emerges during interaction.

Both memory types are essential. Without long-term memory, the agent lacks the necessary background to perform any meaningful reasoning; without working memory, it cannot accurately interpret buyer intent, rendering tool invocation ineffective. Together, they empower the agent to reason with context, act with precision, and adapt dynamically across diverse e-commerce scenarios.

\subsection{Decision-Making Module}

The decision-making module is responsible for generating optimal action plans based on the current system state and memory. To this end, we adopt a ``Propose-Evaluate-Select'' framework, where the agent first generates multiple candidate plans, evaluates them based on their ability to fulfill the buyer intent, and deterministically selects the best one for execution.

Candidate plans may consist of sequences of tool invocations, single-tool calls, or direct responses without tool use. This framework encourages the agent to consider diverse possibilities and avoid shortsighted decisions limited to immediate next steps. By modeling plan selection as a discrete classification task, it further enables reliable confidence estimation, capitalizing on LLMs’ intrinsic calibration properties in single-token output tasks~\cite{achiam2023gpt, qun2024xmodel2technicalreport}. The resulting confidence scores reflect the reliability of each option, supporting more interpretable and trustworthy decision-making.

This structured approach facilitates adaptability to diverse scenarios, transparent evaluation through explicit scoring, and robustness in action selection, significantly improving agent performance in complex and dynamic e-commerce interactions.

\subsection{Action Module}

The action module defines all executable operations in MindFlow, including both external and internal actions. This design enables precise and flexible behavior in dynamic e-commerce environments. External actions refer to the invocation of predefined tools that allow the agent to interact with environment and retrieve task-relevant information. Internal actions, in contrast, involve intra-agent operations such as memory retrieval, internal reasoning steps, and status updates. 

MindFlow incorporates a specialized mechanism inspired by the Agent-Computer Interface (ACI)~\cite{yang2024swe} to simplify input complexity and enable structured, LLM-friendly tool invocation. Specifically, MindFlow adopts an LLM-friendly command abstraction to handle token-heavy inputs frequently encountered in e-commerce scenarios, such as image, product, and order URLs. These long tokens are replaced with compact placeholders (e.g., ``[Image 1]''), which are easier for the LLM to process. At runtime, the placeholders are resolved via a multimodal pipeline that retrieves the original content and generates concise textual descriptions. This mechanism generalizes to various forms of long inputs, including video links and structured metadata.

This strategy substantially reduces token consumption and cognitive load for the LLM, leading to faster inference and improved reasoning accuracy. It also enhances the robustness and generalizability of the action module, allowing MindFlow to handle diverse multimodal tasks in a more scalable and efficient manner.

\subsection{Multimodal Integration Strategy}

Multimodal interactions play an important role in e-commerce customer service, as buyers often upload images alongside text queries to describe issues such as damaged goods, unclear specifications, or installation problems. However, relying on a monolithic multimodal LLM to interpret these inputs often results in verbose, unfocused, or hallucinated responses, especially when visual content lacks clear grounding instructions.

MindFlow addresses this challenge through a modular ``MLLM-as-Tool'' paradigm, where the MLLM is regarded as a callable visual reasoning unit rather than the primary planner. Instead of prompting the MLLM to generate full response, the agent provides targeted instructions (e.g., “Describe the damage shown in the image”) and integrates the resulting descriptions into the broader decision-making process. This design cleanly separates visual perception from high-level reasoning and aligns with MindFlow’s agentic design.

The modular ``MLLM-as-Tool'' strategy further enhances system clarity and facilitates debugging by explicitly decoupling visual perception from high-level planning. Each component generates intermediate outputs that are easy to inspect, allowing developers to trace and audit the agent's reasoning pipeline.

\section{Experiments and Evaluation}

We conduct both online A/B testing and simulation-based ablation study to comprehensively assess MindFlow’s performance in e-commerce customer service scenarios. These complementary approaches validate the MindFlow’s effectiveness in real-world deployments and its robustness under controlled conditions. Additionally, we compare two multimodal integration strategies to examine how the role assignment of MLLM affects task performance.

\subsection{Online A/B Testing}

In the online A/B testing, MindFlow is deployed in a real-world e-commerce store for household essentials, targeting two high-frequency service scenarios: product consultation and logistics \& order support. The system is equipped with four domain-specific tools to support real-time task resolution: product information retrieval, order management, logistics tracking, and a multimodal tool. While the first three tools primarily serve specific scenarios, the multimodal tool operates across diverse scenarios, enabling flexible visual-textual understanding.

To evaluate MindFlow’s performance, we use the AI contribution ratio (defined in Equation~\ref{eqa: AI_contribution_ratio}) as the primary metric, comparing it against the rule-based customer service system. Buyers are randomly assigned to either the experimental group (interacting with MindFlow) or the control group (interacting with the rule-based system).

\begin{equation}
\text{AI Contribution Ratio} = \frac{V_{\text{AI}}}{T_{\text{AI}} + T_{\text{CR}}}
\label{eqa: AI_contribution_ratio}
\end{equation}

where $V_{\text{AI}}$ denotes the number of AI-generated messages judged by customer representatives to be contextually appropriate and relevant within the dialogue. $T_{\text{AI}}$ denotes the total number of AI-generated responses, and $T_{\text{CR}}$ denotes the total number of responses from customer representatives.

Table~\ref{tab:ai-contrib-online} presents the comparative results of AI contribution ratio in online A/B testing across different service scenarios.

\begin{table}[ht]
  \centering
  \renewcommand{\arraystretch}{1.1}
  \setlength{\tabcolsep}{5pt}
  \begin{tabular*}{\linewidth}{@{\extracolsep{\fill}}ccc}
    \toprule
    \textbf{Scenario} & \textbf{Rule-Based} & \textbf{MindFlow}\\
    \midrule
    Product Consultation\xspace&31.39\%&\textbf{89.82\%}  \\
    Logistics \& Order & 64.56\% & \textbf{65.15\%}\\
    \bottomrule
  \end{tabular*}
  \newline
  \caption{AI Contribution Ratio in Online A/B Testing}
  \label{tab:ai-contrib-online}
\end{table}

Table~\ref{tab:ai-contrib-online} shows that MindFlow achieved a substantially higher AI contribution ratio than the rule-based system in the product consultation scenario, with a relative improvement of 186.14\%. This improvement stems from the limitations of the rule-based system, which relies on manually configured responses. Due to incomplete coverage and infrequent updates, it often fails to reflect the latest product specifications, leading to outdated or inaccurate answers. In contrast, MindFlow invokes tools to retrieve real-time product information dynamically, ensuring broader coverage and higher accuracy.

In the logistics \& order support scenario, both systems performed similarly. These queries mainly pertain to fixed, well-specified information like order status or store policies, where rule-based systems already provide robust coverage through explicit configurations, leaving limited room for improvement from dynamic tools.

This online deployment provides a direct and practical evaluation of MindFlow’s performance in real-world customer service settings. On average, MindFlow achieves a 93.53\% relative improvement over the rule-based system across both scenarios.

\subsection{Simulation-Based Ablation}

Although online A/B testing confirms MindFlow’s practical effectiveness, it offers limited visibility into the contribution of individual system modules and covers only a narrow set of service scenarios. To address these limitations, 
we conduct a simulation-based evaluation using ECom-Bench~\cite{ECom-Bench}, a benchmark specifically designed for e-commerce customer service. ECom-Bench consists of 53 manually validated task instances derived from authentic e-commerce dialogues (18 multimodal and 35 unimodal), augmented with human-curated variations to reflect realistic complexity across diverse service types. This simulation environment replicates key workflows such as pre-sales consultation, after-sales support, and complaint handling, offering a controlled and reproducible platform for fine-grained performance analysis. It enables systematic evaluation of MindFlow’s capabilities in handling complex multi-turn interactions, multimodal understanding, and user intent resolution.

System robustness and reliability are evaluated using the pass\textasciicircum k metric (defined in Equation \ref{eqa: pass_hat_k}), which measures the probability that all $k$ independent trials for a given task are successful \cite{yao2024tau}. This metric reflects the stability of the agent’s performance under varying dialogue phrasings but consistent task semantics, capturing success consistency critical for real-world applications requiring rule adherence and high fidelity.

\begin{equation}
\text{pass\textasciicircum k}=\mathbb{E}_{\text{task}}\left[\binom{c}{k} \middle/ \binom{n}{k}\right] 
\label{eqa: pass_hat_k}
\end{equation}

where $n$ denotes the number of trials for a given task, $c$ denotes the number of successful completions, and $k$ denotes the number of trails under evaluation.

We choose $k=5$ in our evaluation to strike a balance between efficiency and computational overhead, providing dependable robustness measurement. For ablation analysis, we isolate the contributions of the decision-making module and ACI within the action module. The memory module is excluded, as it is tightly coupled with both components and fundamental to user intent understanding. Its absence would render most tasks unsolvable and the comparison uninformative. This setup enables controlled evaluation of decision and action logic under realistic task conditions.

\begin{figure}
    \centering
    \includegraphics[width=1\linewidth]{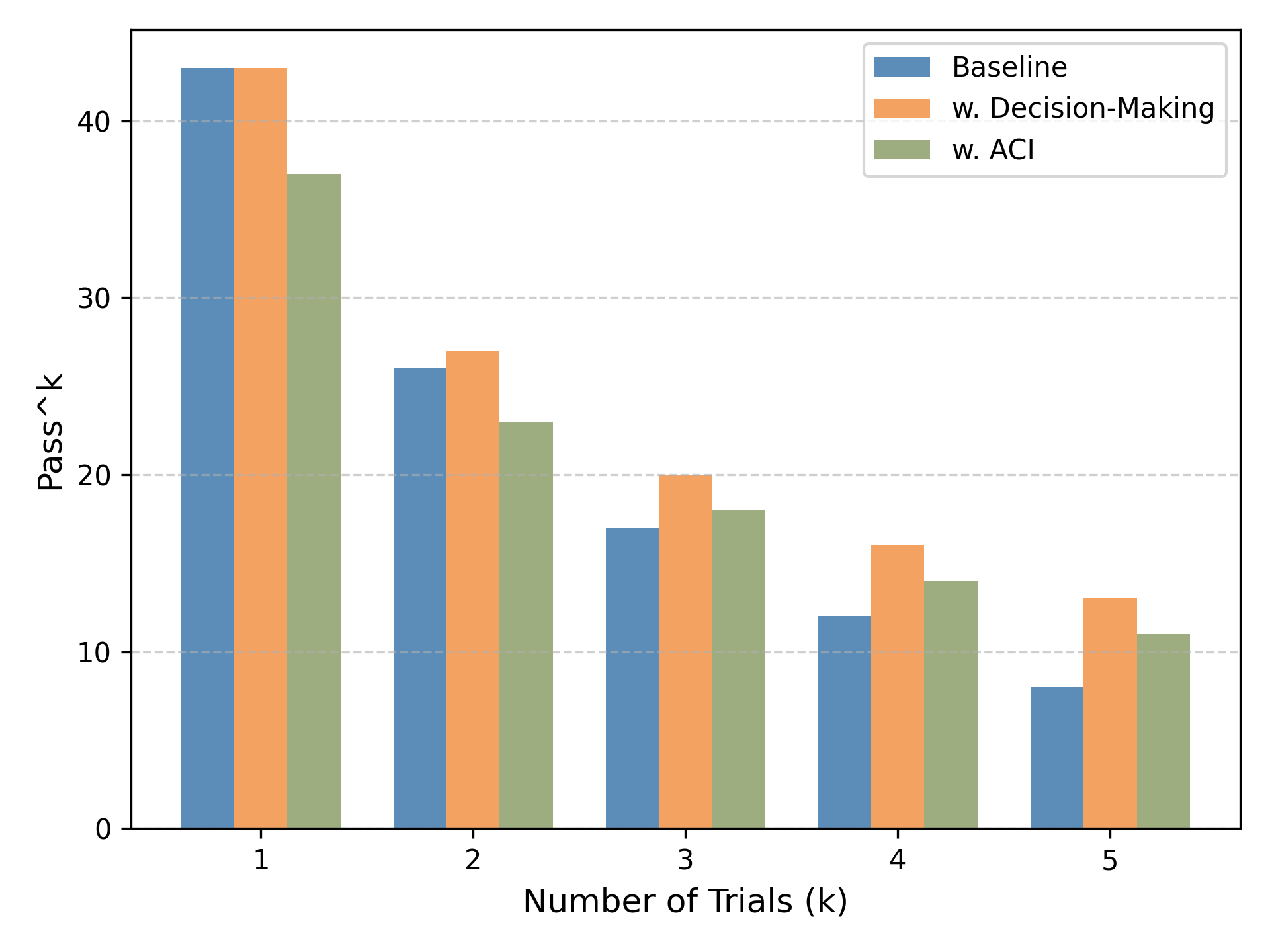}
    \caption{Module Ablation Performance Comparison}
    \label{fig:ecom-bench_results}
\end{figure}

Figure~\ref{fig:ecom-bench_results} presents the module ablation results. At lower values of $k$ (e.g., $k=1$), the performance of both the decision-making module and ACI slightly lags behind the baseline due to increased randomness in few-shot settings. However, as $k$ increases ($k \geq 3$), the effect of stochasticity diminishes, and the benefits of these modules become more apparent. At pass\textasciicircum 5, ACI achieves a 37.5\% relative improvement, while the decision-making module yields an even higher gain of 62.5\%, highlighting their respective contributions to overall system robustness.

The decision-making module’s strong gains stem from its ability to evaluate diverse candidate plans and generate well-calibrated confidence scores, enhancing the reliability of action selection and improving task success rates. Similarly, the ACI mechanism improves robustness and decision accuracy by abstracting lengthy, unstructured links into symbolic tokens, which reduces input complexity, mitigates parsing errors, and promotes more stable and predictable agent behavior.

To further highlight ACI’s efficiency, we measure the average task completion time for both the baseline and the ACI-enhanced baseline, as shown in Table~\ref{tab:ablation-ecombench-aci}. Since the current ACI implementation primarily replaces image URLs with placeholders, time savings in unimodal scenarios are minimal. However, in multimodal tasks, a significant improvement is observed, with task completion time reduced by approximately 48.84\%.

\begin{table}[h]
\centering
\begin{tabular}{ccc}
\hline
\textbf{Configuration} & \textbf{$\bm{\overline{t}}_{\text{multimodal}}$}(s) & \textbf{$\bm{\overline{t}}_{\text{unimodal}}$}(s)\\

\hline
Baseline & 11.59 & 5.39 \\
w. ACI & 5.93& 5.19\\
\hline
\end{tabular}
\caption{Average Task Completion Time Ablation}
\label{tab:ablation-ecombench-aci}
\end{table}

Simulation-based ablation demonstrates that both the decision-making module and the ACI are crucial components driving system performance. The decision-making module enhances accuracy and robustness, while the ACI significantly accelerates response time in multimodal tasks, demonstrating the effectiveness of MindFlow’s modular design.

\subsection{Multimodal Integration Strategy}

We compare two multimodal integration strategies that assign different roles to MLLMs in processing text-image queries to evaluate their impact on e-commerce task performance. The first, ``MLLM-as-Tool'', treats the MLLM as a specialized visual processing tool, while the second, ``MLLM-as-Planner'', uses it as the main planner orchestrating reasoning. For each strategy, we select representative models from the same model family for a fair comparison: the text-only models Doubao-1.5-Pro-32k and Qwen-Max serve as the tool configuration, whereas the multimodal models Doubao-1.5-Vision-Pro-32k and Qwen-VL-Max represent the planner configuration. For simplicity, these models are referred to as ``Doubao'' and ``Qwen'' in the figure legends.

\begin{figure}
    \centering
    \includegraphics[width=1\linewidth]{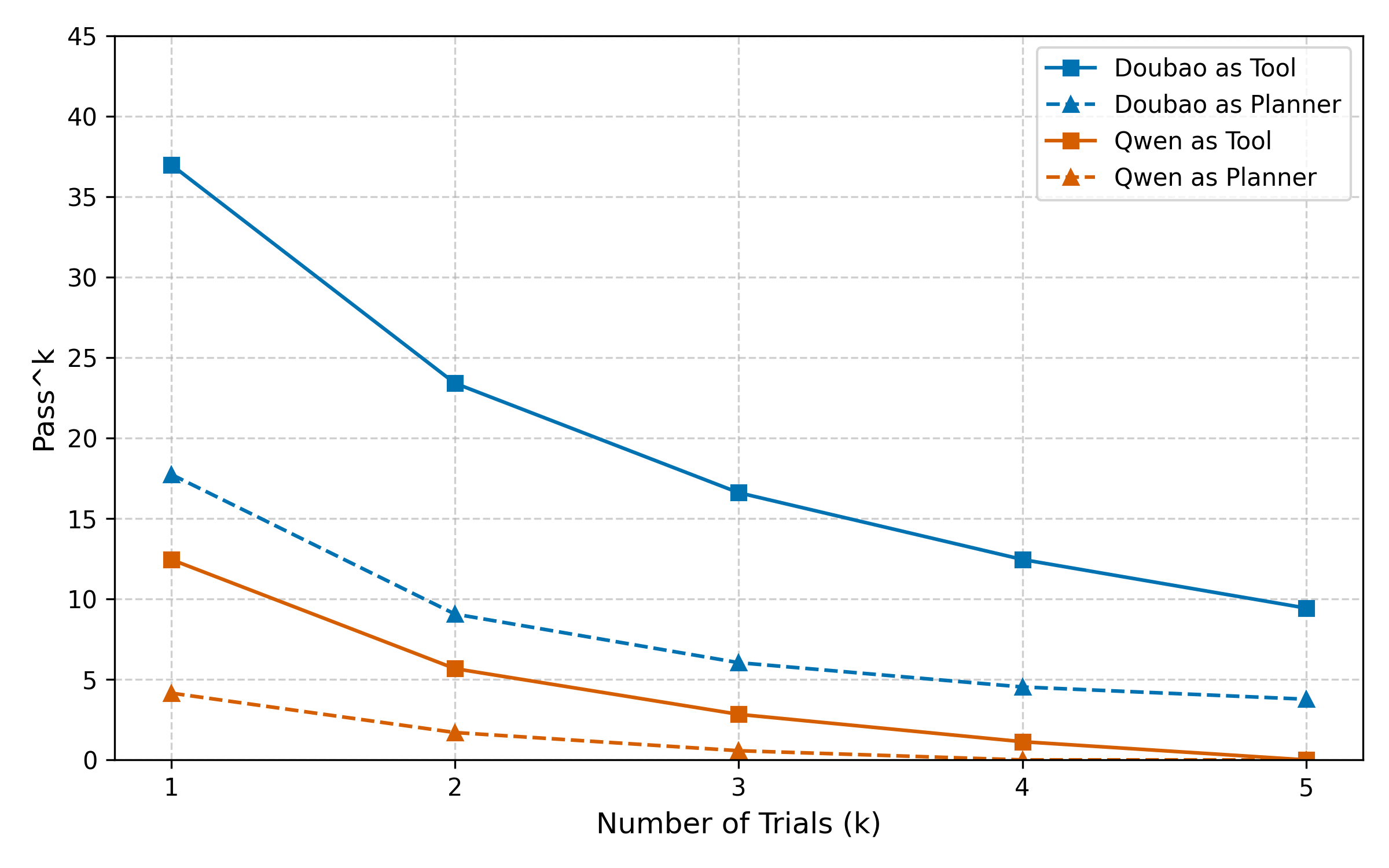}
    \caption{Multimodal Integration Strategy Comparison}
    \label{fig:multimodal_strategy_comparison}
\end{figure}

Figure~\ref{fig:multimodal_strategy_comparison} demonstrates that the ``MLLM-as-Tool'' strategy consistently outperforms the ``MLLM-as-Planner'' approach across all evaluated settings, achieving relative improvements of 108.46\% for Doubao and 200\% for Qwen at pass\textasciicircum1. At pass\textasciicircum5, both Qwen variants (tool and planner) fail to complete any trials, while Doubao shows a 150.13\% relative improvement when using the ``MLLM-as-Tool'' strategy. Furthermore, the performance gap widens as task complexity and model capability increase. Notably, the ``MLLM-as-Planner'' strategy suffers from significantly higher failure rates in visual-semantic alignment, indicating challenges in holistically integrating multimodal inputs.

These results suggest that the modular design inherent in the ``MLLM-as-Tool'' paradigm enhances system stability and robustness by isolating error sources and reducing the complexity of the reasoning process. By delegating visual processing to a specialized module, this approach streamlines the decision-making pipeline, leading to improved accuracy and reliability in multimodal e-commerce tasks.

\section{Conclusion}

MindFlow, the first open-source multimodal LLM agent for e-commerce, is proposed to address the mounting challenges in e-commerce customer service, such as handling complex multimodal queries in real time while maintaining high user satisfaction. Built upon the CoALA framework, MindFlow integrates a memory module for contextual awareness, a decision-making module based on a ``Propose-Evaluate-Select'' framework for adaptive and robust planning, and an action module featuring the ACI for simplifying complex inputs and enhancing interpretability to ensure reliable tool usage. Its core innovation, the ``MLLM-as-Tool'' paradigm, treats multimodal LLMs as specialized visual processors, effectively reducing hallucinations and enhancing overall system robustness.

In real-world online A/B testing, MindFlow achieves a 93.53\% relative improvement over rule-based systems. Simulation-based ablation studies on ECom-Bench further confirm the significant contributions of the decision-making module and the ACI within the action module to overall system robustness. Experiments further confirm the benefits of the ``MLLM-as-Tool'' paradigm in improving multimodal reasoning, contributing to MindFlow’s robust performance.

Building on these promising results, our future efforts will focus on extending MindFlow’s capabilities in several important aspects, such as dynamic long-term memory updates, enhancing the calibration rejection mechanism in the decision-making module for more reliable confidence-based rejection, broadening input abstraction via ACI, and improving resilience against latency and failures in external tool calls. Further evaluation in diverse and high-stakes environments will also be pursued to validate the system’s broader effectiveness.
\section*{Limitations}

While MindFlow demonstrates strong performance, several limitations remain. First, the memory module lacks dynamic long-term memory updates, reducing the agent’s ability to retain evolving user preferences. Second, the decision-making module relies on heuristic confidence scores; improving calibration could enable more principled rejection handling. Third, the ACI currently abstracts only image links, limiting its effectiveness in tasks involving other complex inputs like videos or structured metadata. Finally, system performance depends on external tools such as MLLMs and APIs, which may introduce latency or failure in deployment environments.


\bibliography{main}

\end{document}